\documentclass[default]{sn-jnl}

\hypersetup{
  pdftitle={AU Dataset for Visuo-Haptic Object Recognition for Robots},
  pdfauthor={Bonner, Buhl, Kristensen, and Navarro-Guerrero},
  pdfsubject={Technical Report, Dataset description},%
  pdfkeywords={Multimodal Object Recognition, Data Fusion, Visuo-Haptic Object Recognition, Tactile Perception, Haptic Information, Kinesthetic Perception, Robot Perception},%
  colorlinks=true,
  linktoc=all,
  linkbordercolor=green,
  linkcolor=blue,
  citecolor=darkgray,
  urlcolor=black,
}

\begin{document}

\title[AU Dataset for Visuo-Haptic Object Recognition]{AU Dataset for Visuo-Haptic Object Recognition for Robots}

\author[1]{\fnm{Lasse Emil R.} \sur{Bonner}}

\author[1]{\fnm{Daniel Daugaard} \sur{Buhl}}

\author[1]{\fnm{Kristian} \sur{Kristensen}}

\author*[1]{\fnm{Nicol\'as} \sur{Navarro-Guerrero}}\email{nicolas.navarro.guerrero@gmail.com}

\affil*[1]{\orgdiv{Department of Electrical and Computer Engineering}, \orgname{Aarhus University}, \orgaddress{\street{Finlandsgade 22}, \postcode{8200} \city{Aarhus}, \country{Denmark}}}

\abstract{Multimodal object recognition is still an emerging field. Thus, publicly available datasets are still rare and of small size. This dataset was developed to help fill this void and presents multimodal data for 63 objects with some visual and haptic ambiguity. The dataset contains visual, kinesthetic and tactile (audio/vibrations) data. To completely solve sensory ambiguity, sensory integration/fusion would be required. 
This report describes the creation and structure of the dataset. The first section explains the underlying approach used to capture the visual and haptic properties of the objects. The second section describes the technical aspects (experimental setup) needed for the collection of the data. The third section introduces the objects, while the final section describes the structure and content of the dataset.}


\maketitle
Downloading and Citing this dataset:
Bonner, L.\ E. R., Buhl, D.\ D., Kristensen, K., \& Navarro-Guerrero, N.\ (2021). AU Dataset for Visuo-Haptic Object Recognition for Robots. figshare. \url{https://doi.org/10.6084/m9.figshare.14222486}

\section{Object Exploration}
\label{sec:exploration}
In order to create the dataset, it is necessary to capture both visual and haptic information from the selected objects. When choosing how to explore the objects, the primary consideration is to capture relevant information from both the visual and haptic modalities. Object explorations are based on the \textit{exploratory procedures} described by Lederman and Klatzky \cite{Lederman1987Hand, Lederman2009Haptic}. 
The haptic properties of the objects can be divided into two sub-categories: kinesthetic, which comprises size, shape and weight, and tactile, which comprises texture and hardness. The exploration carried out for each object can be divided into three phases:

\begin{itemize}
    \item Visual object exploration, 
    \item Kinesthetic object exploration, 
    \item Tactile object exploration.
\end{itemize}

\subsection{Visual Exploration}
In this phase, the objects are repositioned three times to expose different faces to the camera. In addition to the three images of the object, one image of only the background is captured, which can be used to implement background subtraction when processing the data further. The three images are taken in optimal lighting conditions where artificial light from multiple angles decreases the shadows from the objects. The resolution of the images is $4640 \times 3472$. Images can later be manipulated by, for instance, scaling, applying noise and other filters to increase the difficulty of the task as needed. Despite the high quality of the images, visuo-haptic integration is still necessary to reliably classify all objects as shown in the thesis of Buhl and Bonner \cite{Buhl2021Biologically}, and Kristensen \cite{Kristensen2021Better}.

\subsection{Kinesthetic exploration}
\label{kinestetic}
For Kinesthetic Exploration, ``Enclosure'' and ``Unsupported holding'' were selected \cite{Lederman1987Hand, Lederman2009Haptic}.

\paragraph{``Unsupported holding''} The object is lifted and held in the hand. This exploratory procedure provides information about the weight of the object. This procedure is implemented by mounting a human-sized robot hand (RH8D \cite{RH8D}) horizontally while placing the object in the hand. By measuring the current needed to keep the hand still, the weight of the objects can be inferred.

\paragraph{``Enclosure''} The object is enveloped with the hand(s) as much as possible. This exploratory procedure provides some information about the global shape of the object. This procedure is implemented by placing the object in the open RH8D hand, then the hand is closed, and the final position of the fingers is recorded.

\subsection{Tactile exploration}
\label{tactile}
For Tactile Exploration, inspiration from the ``Lateral motion'' and ``Pressure'' procedures was taken \cite{Lederman1987Hand, Lederman2009Haptic}.

\paragraph{``Lateral motion'' }
Sideways movements over the object help to extract information about the object's texture. This procedure is implemented by placing the object in the RH8D robotic hand, ensuring that most of the object's surface is exposed. 
Then another robot -- NAO \cite{NAO} -- explores the object by moving one of its fingers across the object in a side-to-side manner. This motion creates vibrations which are recorded and later used to infer the texture of the object. 
This procedure will be referred to as ``feel'' within the dataset. 

\paragraph{``Pressure''}
Two exploratory procedures inspired by the ``Pressure'' procedure described by Lederman and Klatzky are implemented to capture information about the object's hardness. 
The ``Pressure'' procedure consists of applying a force to the object to infer the object's hardness. In this dataset, this procedure recorded data for two implementations. The first implementation, called ``pressure-poke'', uses the NAO robot to poke the object while the object is being held by the RH8D hand. The second implementation called ``pressure-squeeze'', is similar to the implementation of ``Enclosure''. The object is placed in the RH8D hand, and the hand is closed around the object. Then the fingers apply additional force in small increments. The force and number of increments are the same for all objects. 

\section{Experimental Setup}
\begin{figure}[htbp]
    \centering
    \includegraphics[trim={50mm 0 50mm 0}, clip, width=\textwidth]{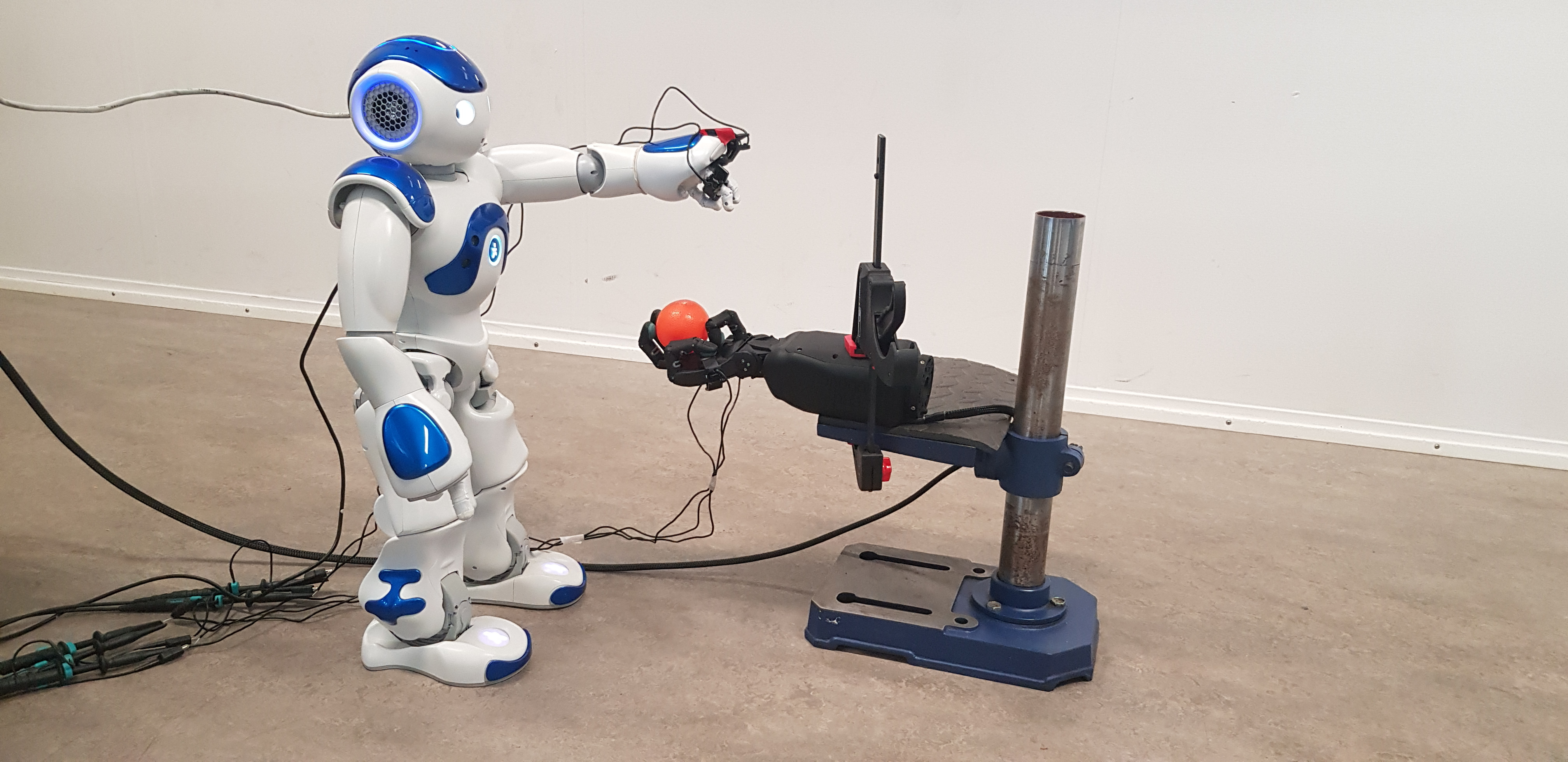}
    \caption{Experimental setup with NAO robot, RH8D Seed Robotics hand and attached microphones.}
    \label{fig:experimental-setup}
\end{figure}

The experimental setup used for the data collection consists of an NAO robot, a human-sized robotic hand RH8D, a camera, an oscilloscope, contact microphones and an array of everyday objects that vary in colour, shape, size, material and weight. Clip-on contact microphones are placed in different locations on the NAO robot and the RH8D hand to measure the vibrations produced when the objects are explored.
The main parts of the experimental setup are shown in Figure \ref{fig:experimental-setup}.

\subsection{Camera}
The camera used is an Olympus OM-D E-M10 Mark III. The camera was mounted on a tripod and at a fixed distance to the objects. The camera was set to auto, and images were taken once the object was in focus. 
The camera remained fixed, and only the objects were repositioned to expose different faces.

\subsection{NAO robot}
The NAO robot used is version 5 with NAOqi version 2.1.14.3. The robot is used to capture the tactile properties (texture and hardness) of the objects by touching and tapping the objects as described in Section \ref{sec:exploration}. 
The NAO robot does not have any built-in sensors that are capable of registering tactile feedback. Thus, five contact microphones are mounted in different locations on both the NAO robot and the RH8D hand. This approach is inspired by the work of Toprak et al.\ \cite{Toprak2018Evaluating}, who showed that tactile sensing is possible using contact microphones in a very similar setup. 

\begin{figure}[htbp]
    \centering
    \includegraphics[width=0.5\linewidth]{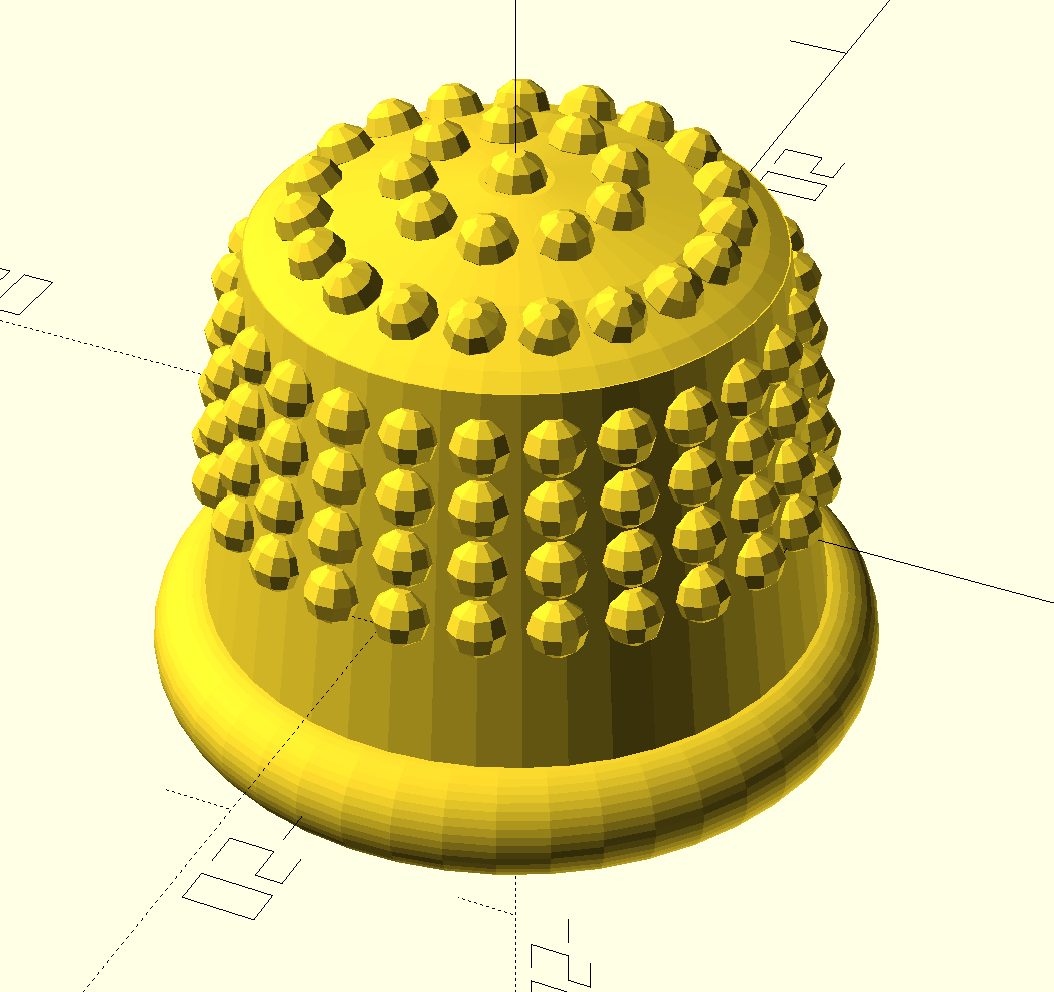}
    \caption{The 3D model used to print the thimble.}
    \label{fig:thimble3d}
\end{figure}

The microphones are used to record the vibrations produced when the robot explores the objects. The robot's fingers are coated with rubber. Thus, the vibrations produced when exploring the surface of the objects are attenuated before reaching the microphones. For this reason, we created a 3D-printed thimble covered with symmetric protuberances, emulating fingerprints, placed on the robot's finger. The 3D model of the thimble can be seen in Figure \ref{fig:thimble3d} and can be downloaded as part of this dataset. 

The thimble is printed in polylactide (PLA) and has a wall thickness of 2mm. The protuberances are 2mm high and distributed around the finger and on the tip with a distance of 0.3mm. The resolution of the thimble is not as fine as a human fingerprint due to the resolution limitations of the 3D printer used, which can print down to a precision of 0.1mm \cite{Ender3}. 
A test was conducted on a few objects from the object suite to ensure that the thimble creates more vibrations than the NAO finger itself. These experiments showed that, at a minimum, using the thimble increased the measured signal by 166\%. On the most significant microphone -- placed right under the object -- the signal strength increased by 470\% using the thimble. Despite the excellent results obtained with this thimble's design, the thimble's design was not systematically optimized. Hence a thorough study of thimbles design and materials is recommended.

Although the NAO robot could also be used to record other haptic data, such as the kinesthetic properties: shape and weight, as it was done by Toprak et al.\ \cite{Toprak2018Evaluating}, the physical size of the robot poses too many constraints on the type of objects that can be used. Additionally, the NAO's hand can only be controlled with a binary signal and does not allow the implementation of the ``pressure-squeeze'' procedure. Thus, the RH8D hand was used for all the kinesthetic exploration procedures and the pressure-squeeze procedure instead.

\subsection{RH8D Seed Robotics hand}
The RH8D hand enabled us to use objects of a broader range of sizes and weight than those possible with the NAO robot. Additionally, the affordability and accessibility of the RH8D hand made it a good option to facilitate reproducibility and potentially expand the dataset. The hand is human-sized, has eight motors, and provides 19 degrees of freedom and 4096 different position readings of each motor \cite{RH8D}. From each actuator, real-time current readings are available and can be used to determine the weight of the objects. All these characteristics make the RH8D Seed Robotics hand a good choice to haptically explore objects.

\subsection{Microphones}
The microphones used are Harley Benton CM-1000 clip-on contact microphones. A PicoScope 4824 with eight channels was used to record the data from the microphones with a sampling rate of 400kHz. A consideration when choosing the sample rate is that the data should be suited for sound source localization as well as object recognition.

To perform localization, it is necessary to distinguish the time of arrival of the vibrations to each of the microphones. The speed of sound in plastic is about 2750 m/s at 20°C \cite{Carlson2003Frequency}. Because the microphones are separated by at least 2cm, it is necessary to sample with at least 275kHz to capture the time difference between them. Moreover, the NAO robot and the RH8D hand are made from different types of plastic and other materials, therefore a sample rate of 400kHz is chosen. 

\subsubsection{Finding the optimal microphone placement}
\label{sec:microphone-placement}
The tactile features, texture and hardness, are reliant on well-placed microphones to capture as many vibrations created when the NAO robot interacts with the object as possible. Multiple configurations were tested to ensure optimal placement of the microphones, which was determined with a three-step procedure. The primary consideration was to increase the difference of the measurements for the different objects -- e.g., a soft rubber ball and a hard wooden block.

Firstly, live measurements were carried out to test promising placements both based on the proximity to the point of contact as well as a decent contact between the microphone and the robot. The best placement candidates are listed in Table \ref{tab:micro-placement-initial}.

\begin{table}[htbp]
    \centering
    \caption{Name and placement of microphones (channel) on the NAO and RH8D. The five best placements are shown in Figure \ref{fig:micro-optimal}.}
    \label{tab:micro-placement-initial}
\begin{tabular}{|l|l|}
\hline
\textbf{Channel}     & \textbf{Placement}   \\ \hline
NP       & NAO palm                      \\ \hline
NF       & NAO finger                    \\ \hline
NW       & NAO wrist                     \\ \hline
R8RU     & RH8D hand, right side, under  \\ \hline
R8LU     & RH8D hand, left side, under   \\ \hline
R8RO     & RH8D hand, right side, over   \\ \hline
R8LO     & RH8D hand, left side, over    \\ \hline
    \end{tabular}
\end{table}

Secondly, all the exploratory procedures were performed on a subset of objects.

\begin{figure}[htbp]
    \centering
    \includegraphics[width=\linewidth]{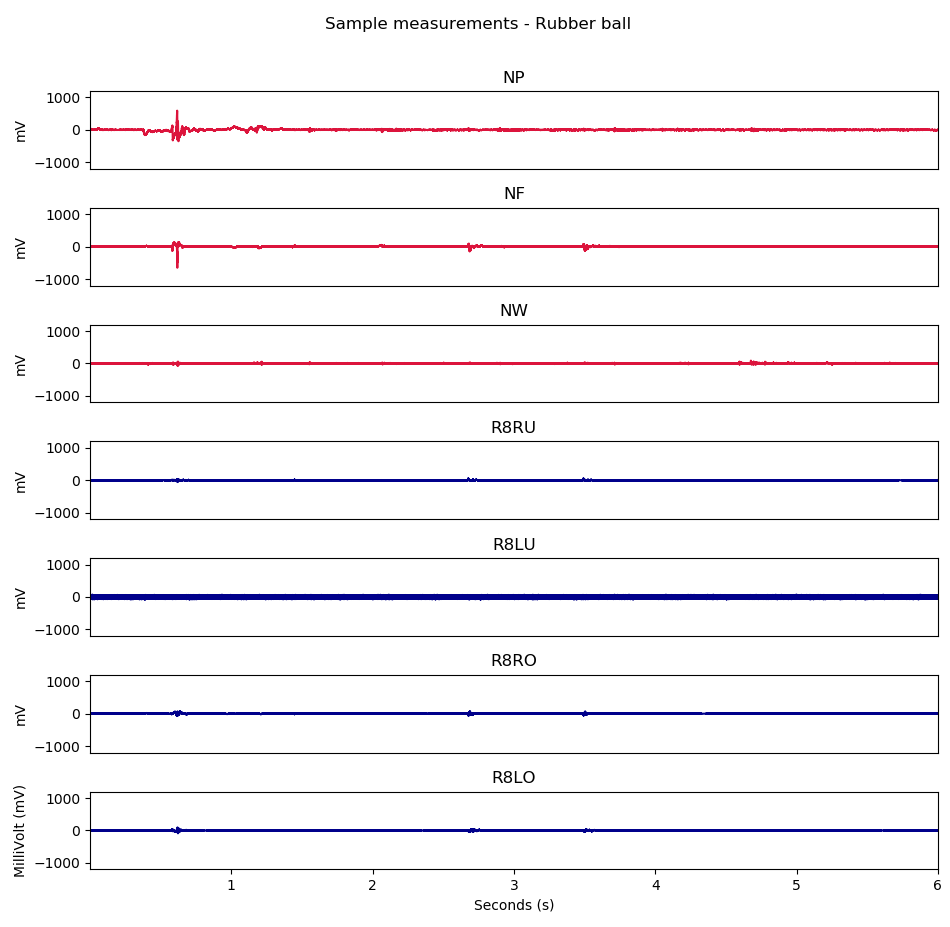}
    \caption{Measurements from all seven microphones when exploring a soft rubber ball. Red is used for Microphones placed on the NAO, and blue is used for microphones placed on the RH8D.}
    \label{fig:ball}
\end{figure}

\begin{figure}[htbp]
    \centering
    \includegraphics[width=\linewidth]{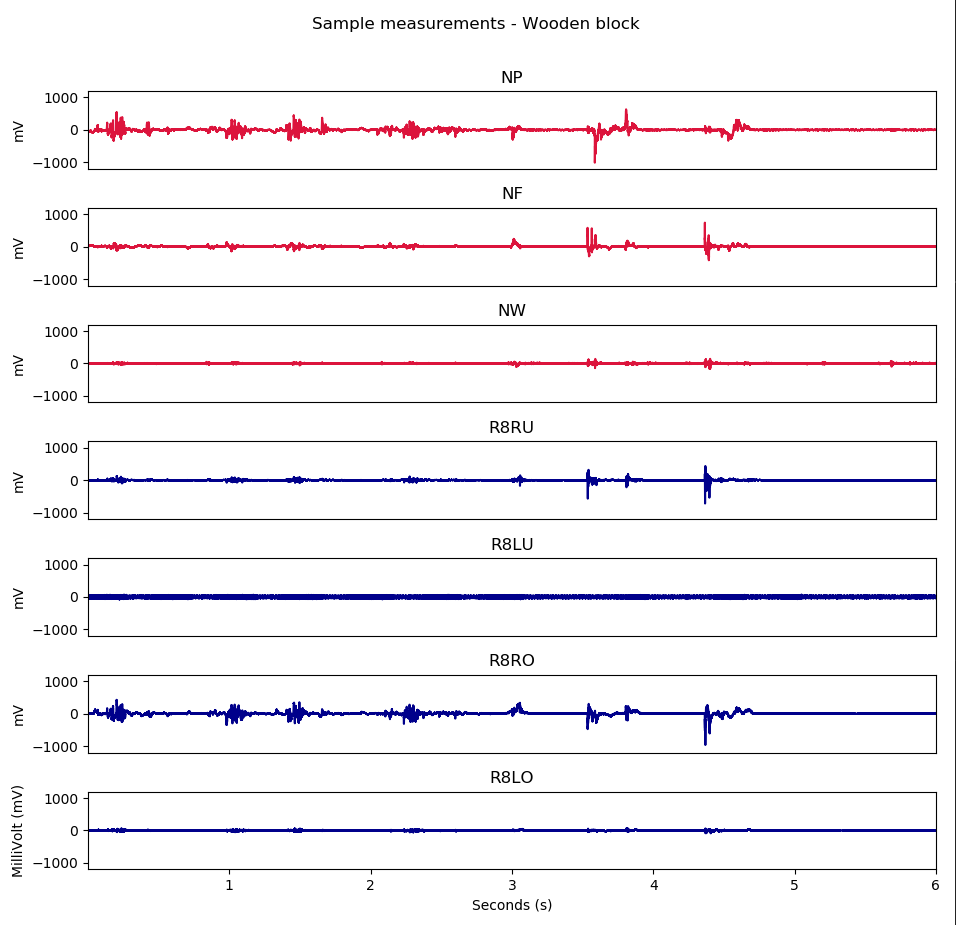}
    \caption{Measurements from all seven microphones when exploring a hard wooden box. Red is used for Microphones placed on the NAO, and blue is used for microphones placed on the RH8D.}
    \label{fig:box}
\end{figure}

Finally, the best placements were determined based on the signal strength. The placements leading to low signal strength were removed from the final setup. 
From the subset of objects, we focused on a soft rubber ball and a hard wooden box as these objects lead to very different signal patterns. Figure \ref{fig:ball} and \ref{fig:box} show the measurements for the lateral motion procedure on a soft rubber ball and a hard wooden box, respectively. 

\begin{figure}[htbp]
    \includegraphics[width=\linewidth]{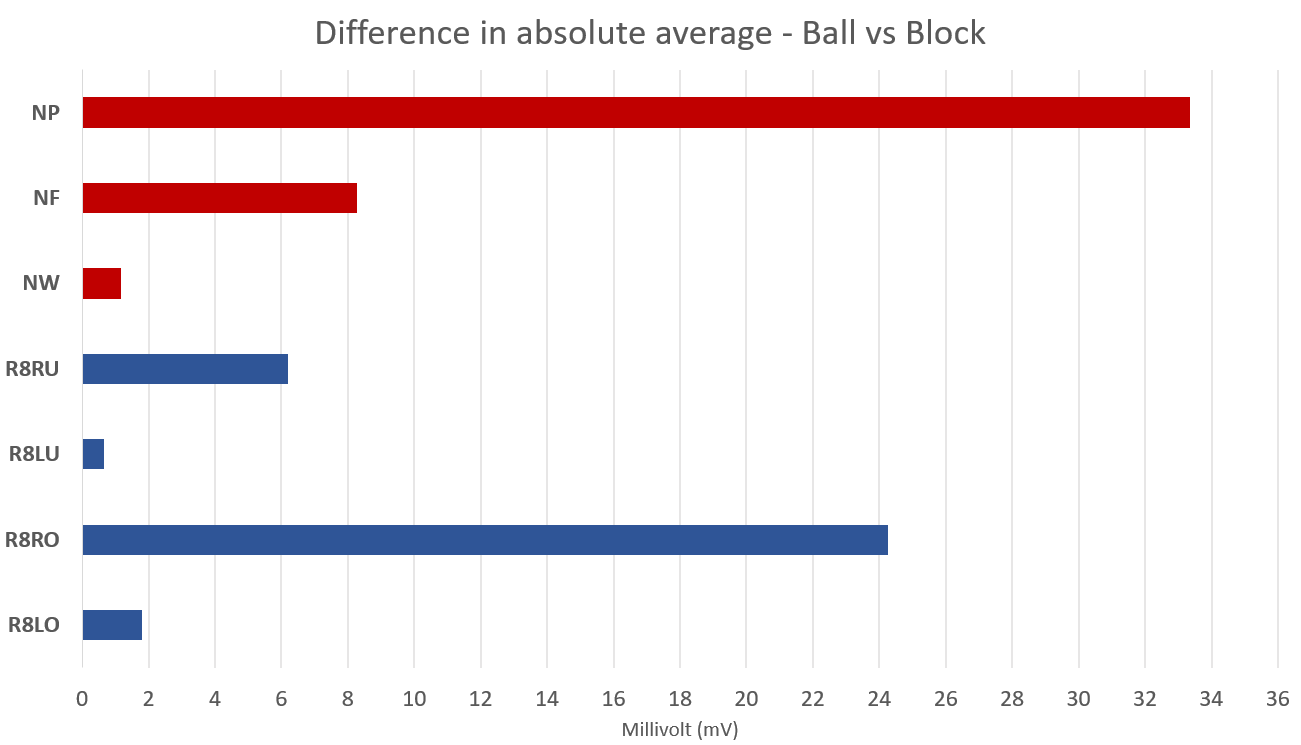}
    \caption{Absolute average difference of the different channels while exploring a wooden box and a rubber ball. Red is used for Microphones placed on the NAO, and blue is used for microphones placed on the RH8D.}
    \label{fig:MPGraph}
\end{figure}

To quantify the difference between the channels, the absolute average difference of a channel is compared across measurements on the two different objects. The differences between the channels are shown in Figure \ref{fig:MPGraph}. The smallest difference is detected in channels NW and R8LU. Thus, these channels are not used in the final setup. The channels NP and R8RO show the most significant difference and are therefore expected to provide a more significant amount of information about the haptic properties of the objects.

\begin{figure}[htbp]
    \includegraphics[width=\linewidth]{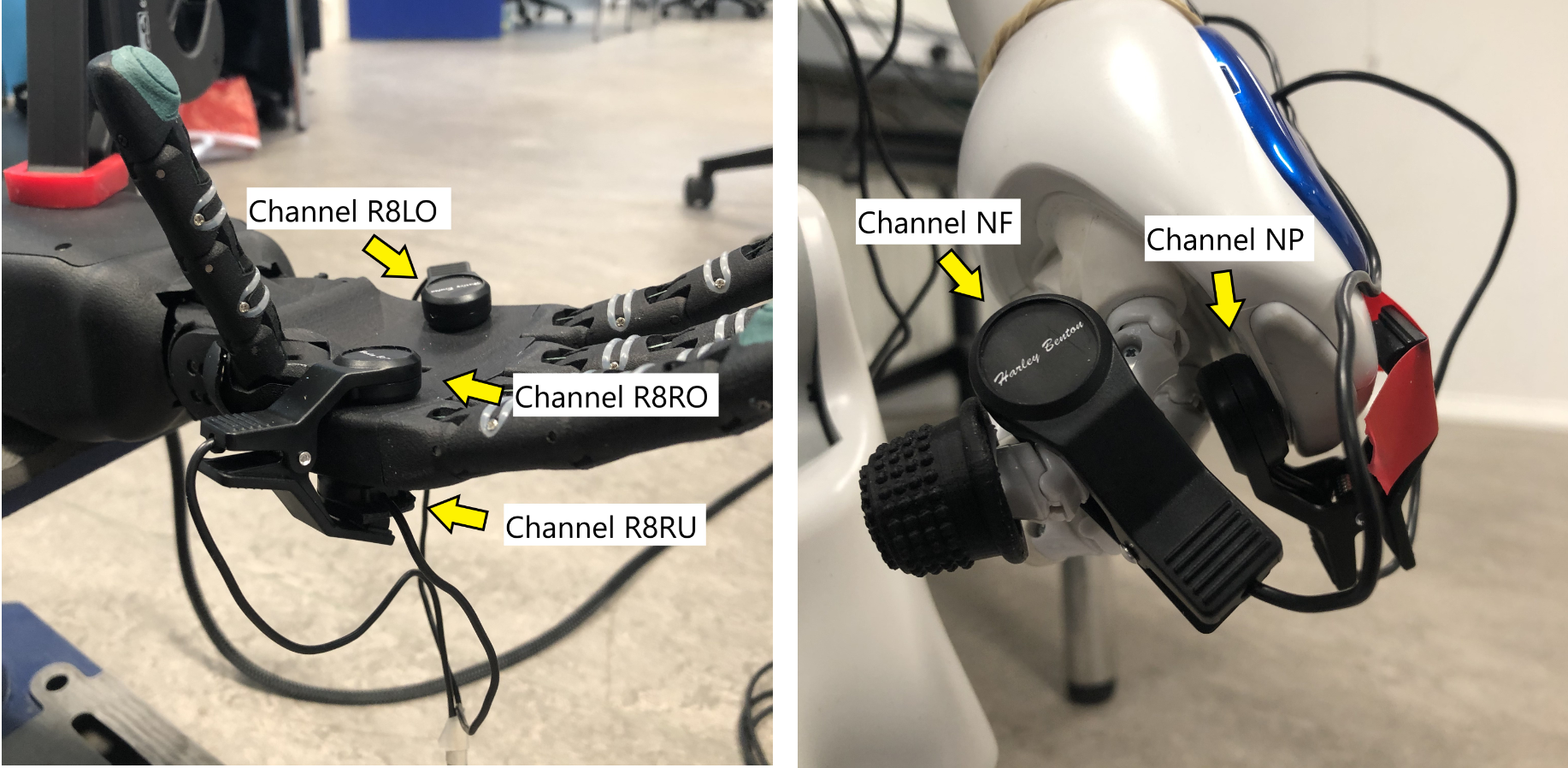}
    \caption{Final microphone setup. Left: the R8HD robotic hand shows channels R8LO, R8RU and R8RO. Right: the hand of the NAO showing channel NF and NP.}
    \label{fig:micro-optimal}
\end{figure}

This leads to the final setup shown in Figure \ref{fig:micro-optimal}, consisting of five microphones from which two are on the NAO and three on the R8HD.


\clearpage
\section{Object Set}
In this section, the considerations used for object selection are presented. Additionally, pictures and properties for each of the objects are listed. 

\subsection{Choosing the objects}
The objects are chosen based on their suitability for multimodal (visuo-haptic) object recognition. We used the following criteria for the dataset:

\begin{itemize}
    \item The dataset should contain visually ambiguous objects
    \item The dataset should contain haptically ambiguous objects
    \item The dataset should contain objects of a variety of materials and colours
    \item The number of objects in the dataset should exceed 50 (most visuo-haptic datasets include less than 50 objects).
    \item The objects should be small enough to fit in the RH8D hand but large enough to facilitate the automated exploration procedures.
\end{itemize}

\begin{figure} [htbp]
    \centering
    \includegraphics[width=1\linewidth]{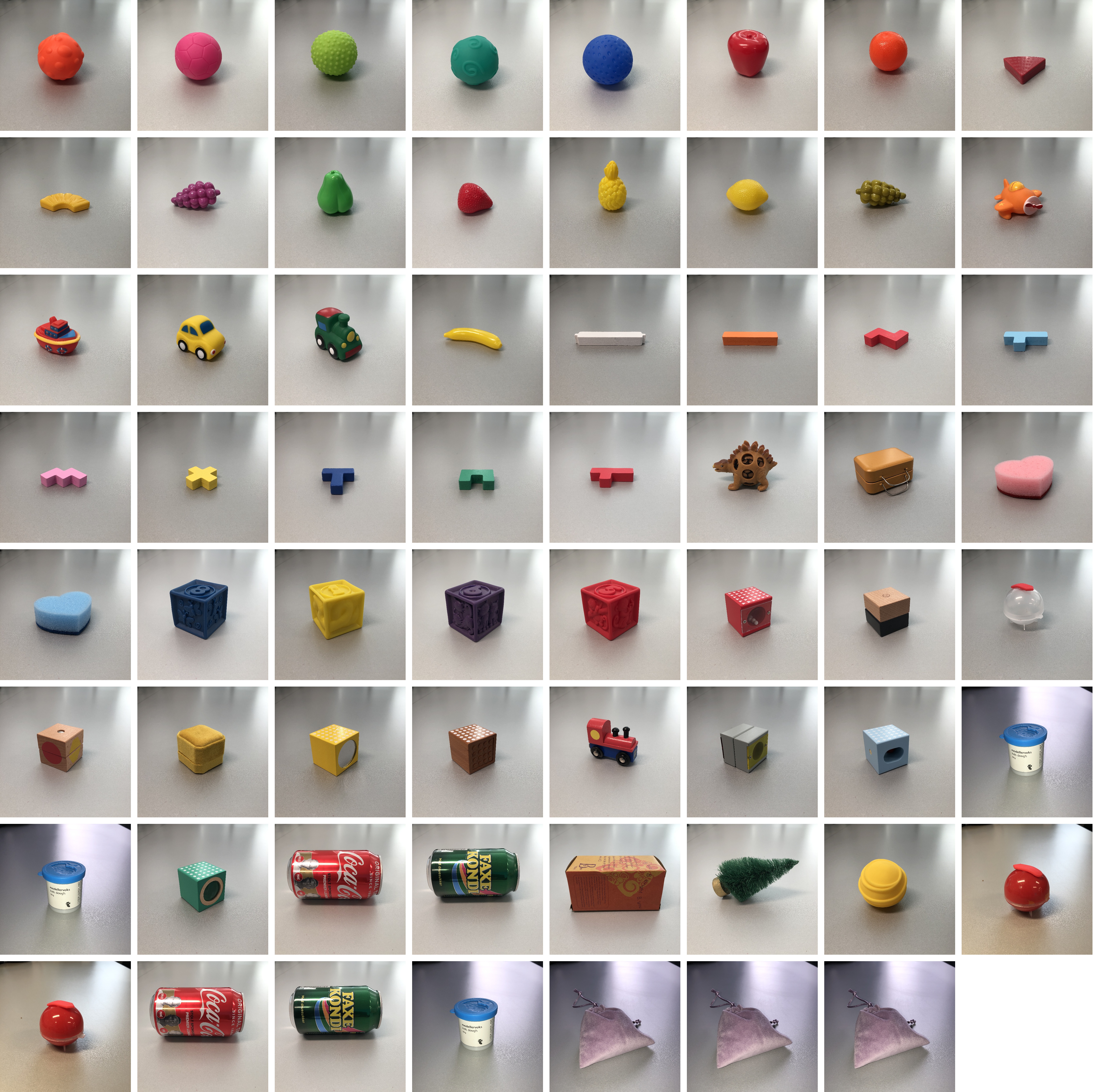}
    \caption{All objects used in the dataset. Some objects are presented several times because they are filled with different substances, which can change the weight and, in some cases, texture and hardness of the object.}
    \label{fig:matrix}
\end{figure}

Figure \ref{fig:matrix} shows an overview of the objects included in the dataset. These images are not part of the dataset but are merely used to identify the objects. 
The objects chosen are a combination of toys and household objects. Especially toys were found to be well suited because they come in different shapes and colours. Objects of similar shape and colour are included to create visual ambiguity. An example of this is the yellow ball and yellow lemon. Some objects are filled with different content, making them only separable by their haptic features. Examples of this are the ice cube container, which is presented as empty and filled with Play-Doh, and the velvet bag filled with 65g salt, coffee beans, and Play-Doh, respectively. Both the ice cube container and velvet bag are presented at the bottom right of Figure \ref{fig:matrix}.
The haptic ambiguity is included by choosing objects of similar material and shape but in a different colour. Examples of this are the balls and cubes. For further specifications of the selected objects, please see the object list uploaded with the dataset. 

\section{Dataset}
The final dataset consists of the previously mentioned visual and haptic data. Each object is measured three times and repositioned each time. This data is collected from three sources:
\begin{itemize}
    \item images of the objects,
    \item kinesthetic data from the RH8D hand,
    \item vibration data from the contact microphones.
\end{itemize}

The visual information consists of four images, a background image for each object and three images of different faces of the object. The distance and heading of the camera remain unchanged for all four images and across objects.

The kinesthetic data collected with the RH8D hand consists of current readings from the wrist and the positions of each of the hand's five fingers. The current from the wrist flexion joint is measured in milliamperes (mA), and the baseline reading (without objects) is 30mA.
The fingers' positions are represented by a value in degrees ranging from -180 degrees (stretched) to 180 degrees (closed). All the information is stored in CSV files. There are individual files for each exploratory procedure (``unsupported holding'', ``enclosure'', and ``pressure-squeeze'') and channels for three different object repositionings. Additionally, readings from the IR proximity sensor (located at the centre of the palm) are provided. All in all, resulting in 60 files for kinesthetic data. 
The current data is stored in the unsupported\_holding\_n.csv files, and the IR proximity sensor reading is stored in the extra\_n.csv files.
The fingers' position data is stored in the enclosure\_n.csv files. 
After the enclosure procedure is sampled, the object is squeezed with four different forces (400, 500, 600 and 700 units of the actuators max force in a 12bit resolution scale, i.e., from 0 to 4095) to investigate the hardness of the objects as described in Section \ref{tactile}. The corresponding fingers' position readings from each force level and the proximity from the IR sensor are stored in the pressure-squeeze\_n.csv files.

The vibration data is recorded in mV with a sampling rate of 400kHz. Immediately after every object's exploratory procedure, the background noise is recorded and stored under the folder called ``background''. The background sample is recorded while the NAO robot is performing exploratory procedures, and the RH8D is in the same position as if it were to hold an object. This is done to compensate in case the noise characteristic from the robots' actuators, cooling fans, and other moving parts changes during data collection.
The vibration data was collected using the five channels/microphones in the positions described in Section \ref{sec:microphone-placement} and stored in CSV files. There are individual files for each exploratory procedure (``feel'' and ``pressure-poke'') and channels for three different object repositionings resulting in 30 files for vibration data. 

The folder structure of the dataset is as follows:

\begin{verbatim}
Data
---|object
------|background
------|observation
---------|haptic_feel_noise
------------|Tactile_feel_n_ch_m.csv
---------|haptic_feel_clean
------------|Tactile_feel_n_ch_m.csv
---------|haptic_pressure-poke_noise
------------|Tactile_pressure-poke_n_ch_m.csv
---------|haptic_pressure-poke_clean
------------|Tactile_pressure-poke_n_ch_m.csv
---------|kinesthetics
------------|enclosure_n.csv
------------|pressure-squeeze_n.csv
------------|unsupported_holding_n.csv
------------|extra_n.csv
------|visual
--------|position-n.jpg
\end{verbatim}

\backmatter

\bibliographystyle{plain}
\bibliography{./references.bib}

\end{document}